\documentclass{article}
\usepackage{spconf,amsmath,graphicx,hyperref}
\usepackage{color}
\usepackage{booktabs}
\usepackage[table]{xcolor}
\usepackage{colortbl}

\title{Exposing the cost of Deep Learning audio development}

\twoauthors
  {Constance Douwes}
  {Centrale Med, Aix Marseille Univ, \\ CNRS, LIS, Marseille, France
  }
    {Paul Magron, Romain Serizel}
	{Université de Lorraine, CNRS\\
	Inria, LORIA, F-54000 Nancy, France}
\begin{document}
\ninept
\maketitle
\begin{abstract}
The environmental impact of deep learning has attracted increasing attention over the past decade. Existing studies mainly focus on the energy and carbon emissions of model training and inference, while the whole development phase is often overlooked. Yet, architecture prototyping and intensive experiments are conducted during this stage, which is highly energy-demanding. In this article, we propose a methodology to estimate these costs, based on activity logs from the Grid5000 shared computing platform used by the LORIA laboratory. As a case-study, we focus on audio projects developed in the Multispeech research team. We evaluate the overall energy cost of four projects, and we compare them to those of training the reported models. Our results show that the energy required for the development phase is 3 to 256 times greater than that required to train the best-performing model alone. These results advocate for a more systematic reporting of energy consumption across the entire life cycle of deep learning-based audio projects.
% and the corresponding carbon emissions can be comparable to those of flying to the conference.
% 164 words <200
\end{abstract}
\section{Introduction}
\label{sec:intro}

Deep learning (DL) has achieved remarkable progress over the past decades, driving advances in numerous scientific domains such as natural language processing, computer vision, and audio signal processing. This progress is following the ever-growing trend to meet the need of a better performance~\cite{amodei2018ai, sevilla2022compute,thompson2020computational}, causing an increase in the associated energy consumption and carbon emissions \cite{strubell2019energy,wu2022sustainable,gupta2021chasing}. In response, an increasing number of studies and tools have been developed to quantify the environmental footprint of DL systems~\cite{Courty2024codecarbon,anthony2020carbontracker,lacoste2019quantifying,lannelongue2021green}, making carbon emissions easier to estimate and report. However, evaluating the environmental impact of the full life-cycle of DL models (i.e., including development, training, and inference) remains under-studied, and existing approaches vary both in the stages of the model life cycle they cover and in the methods used to measure or estimate energy consumption. 

Several studies estimate the carbon emissions of the final model training run. For instance, the GreenMIR study~\cite{holzapfel2024green} estimates the carbon footprint of models based on information reported in papers published at the International Society for Music Information Retrieval (ISMIR) conference. However, only 20\% of the analyzed papers provide sufficient information to perform such an estimation, illustrating the limitations of relying on voluntary reporting. To overcome this issue, organizers of the Detection and Classification of Acoustic Scenes and Events (DCASE) challenge ask authors to measure energy consumption directly during the execution of the final training runs~\cite{Ronchini2022dcasetask4} using the CodeCarbon package~\cite{Courty2024codecarbon}. While direct measurements provide more accurate estimates, it still does not account for the energy costs of the whole development process. This is partly due to the short and well-defined period of time over which the training phase occurs, whereas the development phase is an iterative and exploratory process involving architecture prototyping, repeated trial-and-error, debugging, evaluating, hyper-parameter searches or ablation studies, often spanning over months or years.
%on different setups.  -> pas hyper clair ce que "setup" voulait dire ici (hardware ?) donc je l'ai viré :)

The energy cost of this development phase is rarely quantified unless by the authors themselves when the study specifically aims to track those emissions. One of the pioneering studies in this direction compares the cost of final training with that of the development process of Natural Language Processing (NLP) models~\cite{strubell2019energy}. For two of the studied models, they estimated development costs at 2,000 and 3,000 times the cost of the final training run, respectively. More recent studies have also added the development costs of large-scale language models, although their reported ratios are much smaller. For example, the carbon footprint the development phase of the BLOOM model was estimated at more than twice that of its final training run \cite{luccioni2023estimating}, and the total computational cost of OPT at about twice the training cost as well~\cite{zhang2022opt}. These results illustrate the importance of accounting for the whole development process, while also showing that its relative contribution can vary considerably across studies. Moreover, most of these studies focus on NLP models. In audio and speech processing, analyses of the development phase are still missing, although concerns about the energy and carbon costs of training and inference are growing for speech recognizers~\cite{parcollet21_interspeech}, audio generative models~\cite{douwes2021energy, douwes2023quality}, text-to-speech~\cite{Ronchini2025tts}, and machine listening~\cite{serizel2023performance}, as well as in conferences and challenges such as ISMIR~\cite{holzapfel2024green} and DCASE~\cite{douwes2025energy}.

% In this article, we propose a methodology to estimate the energy consumption of the development phase at different scales, from the \textit{research lab} to the \textit{individual project}. Our methodology leverages computational activity data from Grid5000\footnote{\scriptsize\url{www.grid5000.fr/w/Grid5000}}, a shared computing infrastructure used by French academics. We estimate the energy consumption from the logs of the jobs submitted to the platform, which contain metadata such as the owner, the resources used, the duration and other details about the executed command. We used two different strategies, one based on hardware specifications approximations and the other from direct power measurements. We then aggregate these estimates to obtain the total energy consumed by the lab, the team or by a project.

In this article, we propose a methodology to estimate the energy consumption of the development phase at different scales, from the \textit{research lab} to the \textit{individual project}. Our methodology builds on the logs of the jobs submitted to a shared computing platform, which contain metadata (e.g., owner, resources, duration, executed command, folder) from which we estimate the energy consumption using hardware power specifications and direct power measurements. We apply this methodology to Grid5000,\footnote{\scriptsize\url{www.grid5000.fr/w/Grid5000}} a shared computing infrastructure used by French academics, and conduct a case study at the LORIA\footnote{\scriptsize\url{www.loria.fr/en/loria/}} research laboratory in Nancy, France. We first analyze the long-term trends in the use of accelerators from 2014 to 2025. Then, to focus more specifically on audio research, we measure the daily energy consumption of the Multispeech research team.\footnote{\scriptsize\url{team.inria.fr/multispeech/home/}} Finally, we select four representative projects within this group, and compare their overall energy cost with that required to train the reported models.
Our findings highlight a substantial energy toll at all scales, which advocates for more sustainable audio research practices.
%We found that the energy required for the development phase is 3 to 256 times greater than that required to train the best-performing model alone.
Our code is available online.\footnote{\scriptsize\url{github.com/magronp/g5k_energy}}
% Overall, our findings shows that monitoring development energy consumption advocates for more systematic reporting. 
% While the code is Gris5000 specific, the overall proposed methodology to account for the development energy can be re-used accross. Overall, our findings shows that monitoring development energy consumption advocates for more systematic reporting. 

% The rest of this paper is structured as follows. Section~\ref{sec:methodo} describes our proposed methodology, and Section~\ref{sec:results} presents the results. Finally, Section~\ref{sec:conclu} draws some concluding remarks. % peut être viré pour gagner de la place / mettre un peu plus de biblio.

\section{Methodology}
\label{sec:methodo}

In this section, we present our methodology to estimate the cost of development at different scales. We describe how we estimate the energy consumption of a given task ran on Grid5000, and we then detail the scope and the different scales of our analysis. 

\subsection{Energy consumption}
\label{sec:methodo-energy}

We use two complementary approaches to estimate energy consumption. Both methods operate at the job level, where a \textit{job} refers to any computational task submitted to the scheduler, e.g., short tests, debugging, failed runs, training runs. For each job, we retrieve some metadata such as owner, team, allocated resources (e.g., number and type of node), duration, executed command and launching directory.

The first approach is a static hardware-based estimation method that relies on the green algorithm (GA) framework \cite{lannelongue2021green}. The estimated energy of a job $j$ whose duration is denoted $T_j$ is:

\begin{equation}
E_j = (P_\mathrm{CPU} + P_\mathrm{GPU} + P_\mathrm{MEM})\times T_j,
\end{equation}
where $P_{\mathrm{CPU}}$, $P_{\mathrm{GPU}}$, and $P_{\mathrm{MEM}}$ represents the power of the CPU and GPU and memory respectively. CPU and GPU power are approximated via the thermal design power (TDP) provided by the hardware manufacturer, memory power is estimated using a fixed contribution per allocated memory capacity of 3W per 8Gb, following the GA methodology, and $T_j$ is provided by the job log.

The second approach is based on dynamic power measurements obtained from the baseboard management controller (BMC) at the node level. These measurements are collected through the Grid5000 monitoring tool called \texttt{kwollect}.\footnote{\scriptsize \url{www.grid5000.fr/w/Monitoring_Using_Kwollect}} When several jobs run simultaneously on the same node during a measurement interval $i$, for example when a user allocates only one of several available GPUs on the node, we allocate to each job of interest $j$ a fraction of the measured power as per the following weight:

\begin{equation}
    a_{i,j} =\frac{N^{\mathrm{GPU}}_{i,j}}{\sum_k N^{\mathrm{GPU}}_{i,k}},
\end{equation}
where $N^{\mathrm{GPU}}_{i,k}$ is the number of GPUs allocated to any job $k$, including $j$, running on the same node at the same timestamp.

The energy consumption attributed to job $j$ is then obtained by summing the allocated BMC power over all measurement intervals:
\begin{equation}
E_{j} = \sum_i P_{\mathrm{BMC},i} \times \Delta t \times a_{i,j} ,
\end{equation}
where $P_{\mathrm{BMC},i}$ is the measured power of the machine during interval $i$ and $\Delta t=5$~s is the sampling interval.

Those two approaches are complementary. On the one hand, the GA approach allows to estimate energy consumption quickly over a long period of time, but relies on power approximations. On the other hand, BMC provides direct measurements of the node consumption, but this method is time-consuming and still relies on an approximation due to the job allocation strategy.

Note that since our analysis is conducted retrospectively, we could not estimate the energy for each job using libraries such as CodeCarbon~\cite{Courty2024codecarbon}, since these estimate the energy of a given task while it is running. Nevertheless, we use such an approach to measure the per-epoch energy of several model variants in additional experiments, as detailed in Section~\ref{sec:energy-per-training}.

\subsection{Laboratory and team scale} 

We first analyze the computational activity of the LORIA laboratory on Grid5000. We used the core-hour data available on the intranet for all Nancy clusters, where LORIA is located. We start our analysis in year 2014, when the first GPU server was provided, and extended it to year 2025. In order to account for the growing importance of DL experiments over these years, we distinguish between CPU-only and GPU servers. Since this data is expressed in core-hours and does not report individual jobs information, we apply the static estimation as described above to estimate the corresponding energy consumption and apply it to the core-hour usage.
% \footnote{\scriptsize \url{intranet.grid5000.fr/stats/usage\#nancy-all-20_year}} 
We then analyzed the Multispeech team's daily energy consumption on GPU servers. We gather all the jobs submitted by the members of the team during the studied period. Since jobs can span several days, and since our goal is to monitor closely the consumption, we used the BMC approach. However, as mentioned previously, BMC measurements are time-consuming, therefore we limited our analysis to a 15-month period from September 2023 to December 2024.
%A longer period could have been considered, but we found that it did not provide additional information to our study.   -> je vois toujours pas ce que ça veut dire, vu qu'on l'a pas fait :)

\subsection{User and project scale}  

\subsubsection{Project description}
\label{sec:project-desc}

To select specific projects, we considered users working on diverse subjects, and among the team's top consumers of Grid5000 resources. We identified the following four projects they had contributed to over the past years.
%We collaborated closely with three members of the team, each working on diverse subjects and among the team's top consumers of Grid5000 resources. We asked them to identify projects they had contributed to over the past years. This process led us to selecting the following four projects:

Cui et al.~\cite{cui2023end} proposes an end-to-end multichannel automatic speech recognition system that integrates speaker identity cues. Four models were considered (two baselines and two proposed ones), trained using different chunk sizes, yielding a total of 16 reported training runs. 
%with 3 different chucksizes =12 models ? I counted 16. 

Ayilo et al.~\cite{ayilo2024diffusion} focus on diffusion-based generative models for speech enhancement. The authors propose adding a supervised loss component to the classical unsupervised training objective of diffusion models. They compare the performance of three model training variants, using either the unsupervised or supervised loss, or their proposed combination. These models are evaluated on two different datasets, resulting in a total of 6 reported training runs.

Ayilo et al. 2025~\cite{ayilo2025diffusion} further integrate video conditioning into an audio-only diffusion model~\cite{nortier2023unsupervised} in a fully unsupervised setting, along with a new inference strategy to accelerate clean speech estimation. The study includes the audio-only baseline, a state-of-the-art supervised audio-visual model, and the proposed fully unsupervised audio-visual diffusion model, thus 3 reported training runs.

Lastly, Magron et al.~\cite{magron2026bsrnn} aims to reproduce the system and results from the paper titled ``Band-split RNN for music source separation”~\cite{Luo2023bsrnn}, a popular model in the separation community that is unfortunately not straightforward to replicate. Reproducing the whole pipeline required substantial implementation and experimental effort, and 14 configurations were explored for each of the four target instruments, leading to a total of 56 reported training runs.
%Among these configurations, one outperformed the original results which represents four proposed models (one per instrument).

\subsubsection{From users to project}

For each user, we extracted all the jobs submitted to Grid5000 over the period of development, along with the detailed associated metadata. Then, in close collaboration with the authors, we conducted an annotation process based on this metadata, especially the launching directories and the commands they executed. This allowed to allocate the jobs corresponding to a given project specifically, rather than considering all jobs from that user, since in practice a user conducts various projects in parallel.

Once the jobs per projects are identified, we retrieve their energy consumption using both the GA and BMC methods described in Section~\ref{sec:methodo-energy}. We then sum over all the jobs to compute the total energy footprint of the development phase for each project.
 
\subsubsection{Energy per training}
\label{sec:energy-per-training}

To put the development stage energy in perspective, we compare it with the energy required for model training, which can be done in two ways. On the one hand, we consider the energy used for training all model variants reported in the paper, i.e., the so-called ``reported training runs" described in \ref{sec:project-desc}, and hereafter denoted as ``all". On the other hand, we use as reference the energy required to train the model that achieved the best performance, herein denoted as ``best".

Although those experiments ran on Grid5000, it was not possible to match them to specific job logs. Script names were often reused across different runs, making it difficult to distinguish between experiments. Hence, we retrained some of the reported models in the papers for a few epochs (between 2 and 5) to estimate their energy consumption per epoch, and extrapolate the total energy consumption per run based on the total number of epochs for each model. Note that we carefully avoided redundant re-training for experiments that differed only by the total number of epochs, thereby limiting unnecessary energy consumption.

\subsection{Carbon emissions}
\label{sec:carbon}

To estimate the total carbon emissions of each project, we multiplied its energy consumption by the power usage effectiveness (PUE) of the Grid5000 facility, equal to 1.5, and by the average annual French electricity emission factor of the corresponding development years. To illustrate the effect of electricity carbon intensity, we also consider a hypothetical scenario using the US electricity mix. Electricity emission factors are extracted from the ElectricityMaps tool.\footnote{\scriptsize \url{https://app.electricitymaps.com/}}

As an additional illustrative experiment, we estimate the emissions of a round-trip flight from Paris to the location of the conference where each project was published, using the MyClimate flight calculator.\footnote{\scriptsize\url{https://co2.myclimate.org/fr/flight_calculators/new}} Note that this experiment is conducted for an illustrative purpose, rather than for obtaining a reliable and comprehensive estimate of the authors' actual travel emissions.

\section{Results}
\label{sec:results}

\subsection{Laboratory and team energy consumption}

\begin{figure}
    \centering
    \includegraphics[width=0.9\linewidth]{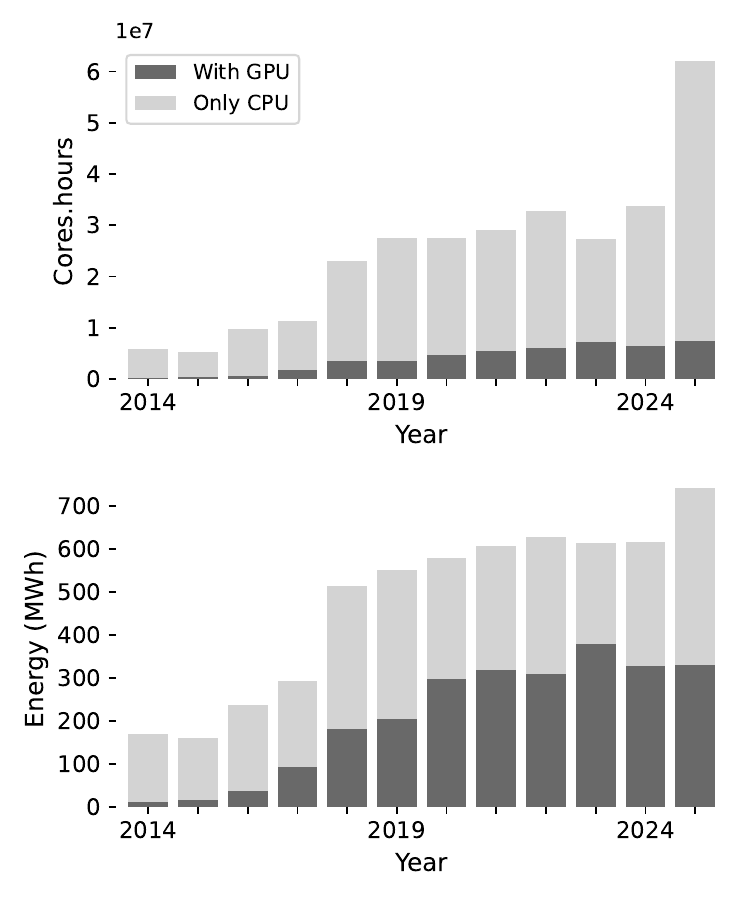}
    \vspace{-1em}
    \caption{Evolution of computational demand (top) and energy consumption (bottom) from 2014 to 2025 for Grid5000 Nancy clusters, with a distinction between CPU-only and GPU servers.}
    \label{fig:gpu_core_energy}
    \vspace{-1em}
\end{figure}

We plot the evolution of computational activity of the LORIA laboratory on the Grid5000 infrastructure in Figure~\ref{fig:gpu_core_energy}. We see that energy consumption increased from 169~MWh in 2014 to 741~MWh in 2025. In 2014, the computational activity and energy consumption were mainly on CPU servers, as only one GPU server was available. This period also corresponds to the early development of DL activities within the laboratory. Over the years, the use of GPU servers increased alongside the deployment of additional GPU servers on the infrastructure. CPU usage also grew, albeit at a slower rate. In the last four years, GPU activity accounted for about 20\% of the allocated core-hours, but represented more than half of the estimated energy consumption. This difference is mainly due to the higher power consumption of GPU servers compared with CPU-only servers. Overall, this analysis highlights the shift towards more energy intensive laboratory practices.

\begin{figure*}
    \centering
    \includegraphics[width=0.9\linewidth]{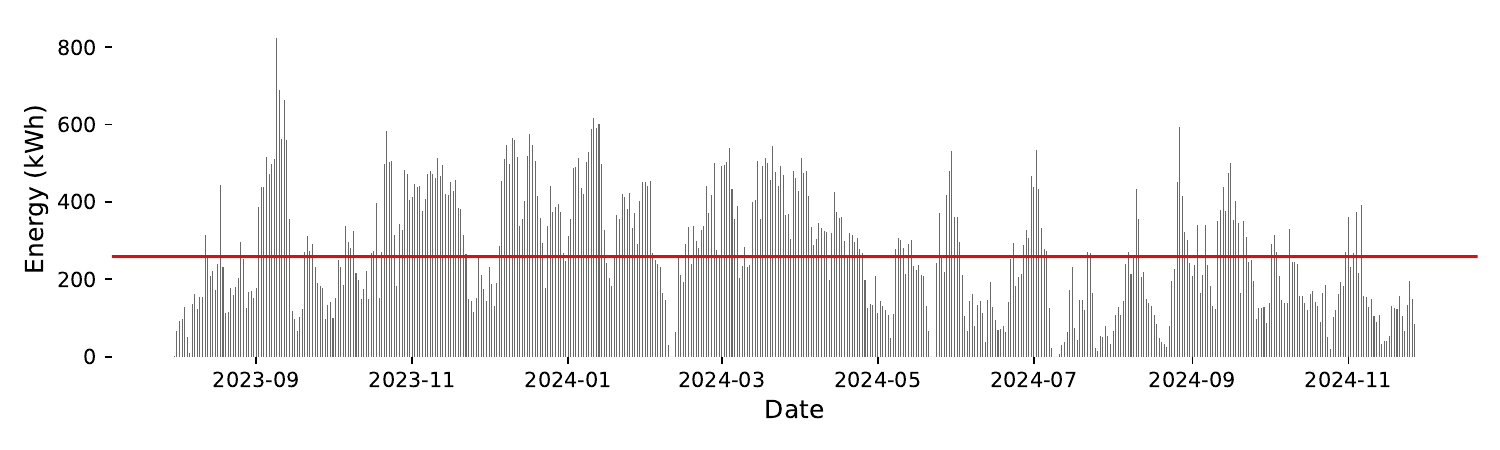}
    \vspace{-1.5em}
    \caption{Daily electricity consumption on GPU servers of Multispeech research group on Grid5000 during the 2023--2024 academic year. The red line indicates the mean daily consumption of 260~kWh.}
    \label{fig:multispeech_profile}
    \vspace{-1em}
\end{figure*}

To examine this evolution at a finer scale and more specifically for audio project development, we plot the Multispeech team's daily energy consumption in Figure~\ref{fig:multispeech_profile}. On average, the team consumes about 260~kWh per day, corresponding to roughly 95~MWh per year. This represents almost one sixth of the total Nancy cluster annual consumption, making Multispeech one of the largest consumers of the platform. To put things
in perspective, the team consumes as much electricity in eight days as an average
French inhabitant does in a year.\footnote{\scriptsize\url{https://ec.europa.eu/eurostat/statistics-explained/index.php?title=Electricity_and_heat_statistics}}
This consumption varies significantly over days, with higher values during periods of intensive development. As team members work on similar topics, periods of increased activity often coincide with paper submission deadlines. Consequently, the platform must have sufficient computing capacity to handle these peak activities, even if these periods are intermittent and difficult to predict. 
% Note that it is a shared infrastructure used by multiple teams whose development cycles and deadlines overlap.
%I don't know how many teams use Nancy site. 

\subsection{Per-project energy consumption comparisons}

\begin{table*}[t]
    \centering
    \caption{Computational activity, energy consumption, and carbon emissions of the four studied projects. ``Jobs" and ``Runs" are the total number of jobs, and reported training runs, respectively. ``Best", ``All", and ``Dev" refer to the estimated energy for training the best performing model, all reported training runs, and the whole project development, respectively. Carbon emissions are estimated under French and US electricity mixes, and air travel corresponds to a round-trip flight between Paris and the conference location.}
    \begin{tabular}{l|rc|rrr|cc|ccc}
        \toprule
        & \multicolumn{2}{c|}{\textbf{Number}} & \multicolumn{3}{c|}{\textbf{Energy (kWh)}} & \multicolumn{2}{c|}{\textbf{Energy Ratio}} & \multicolumn{3}{c}{\textbf{Carbon emissions (tCO$_2$e)}}
        \\
        & \textbf{Jobs} & \textbf{Runs} &
        \textbf{Best} & \textbf{All} &
        \textbf{Dev} & \textbf{Dev/All}& \textbf{Dev/Best} & \textbf{France} &  \textbf{US} & \textbf{Air travel}  \\
        \midrule
        Cui et al. 2023~\cite{cui2023end} & 1,462 & 16 & 11,654 & 17,042 & 35,288 & 2 & \textbf{3} & 2.8 & 22.1 & 4.0 \\
        Ayilo et al. 2024~\cite{ayilo2024diffusion} & 337 & 6  & 222    & 624    & 4,095  & 7 &  \textbf{18} & 0.3 & 2.6  & 4.8 \\
        Ayilo et al. 2025~\cite{ayilo2025diffusion} & 3,529 & 3  & 37     & 123 & 9,462  & 77 & \textbf{256} & 0.5 & 5.9  & 2.8\\
        Magron et al. 2027~\cite{magron2026bsrnn} & 1,463 & 56 & 456  & 3,027  & 24,239 & 8 & \textbf{53} &  1.0 & 12.1 & 2.2 \\
        \bottomrule
    \end{tabular}
    \label{tab:project_energy}
    \vspace{-1em}
    %{\footnotesize \textsuperscript{*}Hypothetical scenario using a US electricity mix. }
\end{table*}

We report in Table~\ref{tab:project_energy} the computational activity, energy consumption, and carbon emissions of the four projects described in Section~\ref{sec:project-desc}. Overall, the number of jobs associated with each project is far larger than the number of reported training runs, as expected given the diversity of tasks involved in the development phase. We also observe that the energy consumed during development is between $3$ and $256$ times that required to train the best performing model, and between $2$ and $77$ times that required to train all reported models. This means that considering only the energy consumption of the reported models, which is common practice, substantially underestimates the energy cost of the whole project.

Looking at each project in more detail, Ayilo et al.~\cite{ayilo2025diffusion} reports only training runs, with a total training energy of 123~kWh of which 37~kWh corresponds to the best model, whereas more than three thousands jobs were executed, consuming a total of 9.5~MWh. This project has therefore the highest development ratios. This may be explained by the publication history of this work: the study was initially submitted to Interspeech and later revised and submitted to ICASSP, with additional rounds of experiments and evaluations. In contrast, the previous project by Ayilo et al.~\cite{ayilo2024diffusion} was submitted only once. It involved 337 jobs that consumed 4.1~MWh, while the 6 reported training runs consumed an 18-th of that total toll. 

For Magron et al.~\cite{magron2026bsrnn}, the number of jobs is around 26 times the number of reported training runs, which is lower than the other projects. One possible explanation is that the project focuses on reproducing an existing pipeline~\cite{Luo2023bsrnn} rather than developing a new one from scratch. Besides, several prototyping or debugging jobs were conducted for a single target instrument only, rather than systematically for all four target instruments. Nevertheless, the replication process still consumed 24.2MWh, which is 53 times the energy required to train the best-performing model. 

Cui et al.~\cite{cui2023end} has the largest absolute energy consumption among the considered projects, for the development phase as well as for training best and all reported models. However the ratio between the development and reported models is the lowest. This is probably due to an incomplete jobs recording of the whole development, since some experiments were ran on Jean Zay, another French computing platform. This means that even though the ratio is low, the total energy is actually underestimated, since these Jean Zay jobs' energy could not be included here. This illustrates an additional difficulty of reporting the total energy cost, considering the diversity of infrastructures that can be used within a single project development.

In terms of carbon emissions, conducting the development process in a French infrastructure has a smaller carbon footprint than traveling to the conference, though not by a large margin. It ranges from about half the travel emissions for Cui et al.~\cite{cui2023end} and Magron et al.~\cite{magron2026bsrnn} to about a tenth of these emissions for Ayilo et al. projects~\cite{ayilo2024diffusion,ayilo2025diffusion}. Considering a hypothetical scenario under the US mix, development emissions far exceed travel emissions for most projects, which highlights the importance of considering the development phase when evaluating a project’s environmental impact.

\subsection{Discussion}
\label{sec:discussion}

Even though this study is conducted retrospectively to the project development, and thus rely on approximations, it provides a first view of the magnitude of the energy and carbon costs of developing DL audio models. While our methodology is applicable to any research project using cloud computing platforms, the released implementation is specific to Grid5000. Finally, the carbon footprint estimations only considers operational emissions from electricity consumption: as such, it does not include embodied hardware emissions, and does not take into account all the other environmental impacts, such as water and resource use, which we leave to future work for refining our methodology.
%on peut aussi dire qu'on a pas prit en compte l'énergie de "laisser allumer" les serveurs ---- ou celle de l'ia générative éclatée au sol pour écrire notre code et nos papiers :D Oh wait

\section{Conclusion}
\label{sec:conclu}

In this work, we proposed a methodology for estimating the energy consumed during the development phase of DL-based audio projects, which we applied at the laboratory, research team, and individual project levels.
%We found that the energy demand of the LORIA laboratory increased from $169$~MWh in 2014 to $741$~MWh in 2025, with GPU clusters accounting for 50\% of total energy consumption. At the team level, the Multispeech research group consumes on average $260$~kWh per day on GPU servers, with a high day-to-day variability. The considered projects required between $3$ and $256$ times more energy during development than to train their best reported model. For some of these papers, the associated carbon emissions are of the same order of magnitude as those of a flight to the conference.
We found both a large general energy increase over time at the lab level, and a substantial energy toll for the audio-oriented Multispeech team only. We also observed that training a single best-reported model often represent only a fraction of the total energy cost of the whole project development. To foster more sustainable practices, we encourage our colleagues to adopt such a methodology, and to monitor energy consumption during the whole life cycle of their projects.


\section{Acknowledgment}
\label{sec:ack}

All computation were carried out using the Grid5000 testbed, supported by a French scientific interest group hosted by Inria and including CNRS, RENATER and several Universities as well as other organizations. The authors would like to thank J.-E. Ayilo and C. Cui for their help in analyzing jobs related to their projects. % moche, à reprendre

\bibliographystyle{IEEEtrans}
\bibliography{references}

\end{document}